%% file: main.tex
\pdfoutput=1

\documentclass[11pt]{article}

\usepackage{acl}

\usepackage{times}
\usepackage{latexsym}

\usepackage[T1]{fontenc}

\usepackage[utf8]{inputenc}

\usepackage{microtype}

\usepackage{inconsolata}

\input{math_commands.tex}

\usepackage{hyperref}
\usepackage{url}

\usepackage[normalem]{ulem}
\usepackage{caption}
\usepackage{subcaption}
\usepackage{graphicx}
\usepackage{wrapfig}
\usepackage{amsthm}
\usepackage{amsmath}
\usepackage{multirow}
\newtheorem{assumption}{Assumption}
\newtheorem{theorem}{Theorem}

\newtheorem{observation}{Observation}

\usepackage{longtable}
\usepackage{booktabs}
\usepackage{xspace}
\usepackage{tabularray}
\usepackage{tabularx}
\usepackage[export]{adjustbox}
\usepackage{xcolor}
\definecolor{red}{rgb}{0.99, 0.02, 0.02}

\NewDocumentCommand{\heng}{ mO{} }{\textcolor{red}{\textsuperscript{\textit{Heng}}\textsf{\textbf{\small[#1]}}}}
\NewDocumentCommand{\chihan}{ mO{} }{\textcolor{blue}{\textsuperscript{\textit{Chi}}{\small[#1]}}}

\newcommand{\Phenomenon}{Position Generalization\xspace}
\newcommand{\phenomenon}{position generalization\xspace}

\title{
Computation Mechanism Behind LLM Position Generalization \\
}

\author{
Chi Han, Heng Ji \\
University of Illinois Urbana-Champaign\\
\texttt{\{chihan3, hengji\}@illinois.edu} \\
}

\begin{document}

\maketitle

\begin{abstract}
    \input{body/0_abstract}
\end{abstract}

\input{figures/teaser}

\input{figures/logit_map}

\input{body/1_introduction}

\input{body/2_related}

\input{figures/logit_disentangle}
\input{body/3_disentangle}
\input{body/3-1_real_attention}
\input{figures/rotating_patterns}
\input{body/3-2_attention_logits}
\input{figures/position_perturbation_ppl}
\input{body/3-4_feature_pattern}

\input{tables/qasper}
\input{body/4_robustness}
\input{tables/length_feature_kl}
\input{body/4-1_order}

\input{figures/sliding_window}

\input{body/4-2_length}
\input{body/5_conclusions}

\input{body/x-1_limitations}
\input{body/x-2_acknowledgement}

\bibliography{custom}

\appendix
\clearpage
\input{body/y-1_ternary_approximation}
\input{body/y-2_fake_logit_approximation}
\input{body/y-3_pattern_mechanism}
\newpage
\input{body/y-5_more_examples}
\newpage
\input{body/y-6_non_triviality}

\end{document}

%% file: math_commands.tex
\usepackage{amsmath,amsfonts,bm}

\def\eqref#1{equation~\ref{#1}}

\def\1{\bm{1}}

\def\va{{\bm{a}}}
\def\vb{{\bm{b}}}
\def\vc{{\bm{c}}}

\def\vk{{\bm{k}}}

\def\vo{{\bm{o}}}

\def\vq{{\bm{q}}}

\def\vv{{\bm{v}}}

\DeclareMathAlphabet{\mathsfit}{\encodingdefault}{\sfdefault}{m}{sl}
\SetMathAlphabet{\mathsfit}{bold}{\encodingdefault}{\sfdefault}{bx}{n}

\def\gR{{\mathcal{R}}}

\def\sR{{\mathbb{R}}}

\newcommand{\E}{\mathbb{E}}



%% file: body/0_abstract.tex
Most written natural languages are composed of sequences of words and sentences.
Similar to humans, large language models (LLMs) exhibit flexibility in handling textual positions - a phenomenon we term \textbf{\phenomenon{}}. They can understand texts with position perturbations and generalize to longer texts than those encountered during training with the latest techniques.
These phenomena suggest that LLMs handle positions tolerantly, but how LLMs computationally process positional relevance remains largely unexplored. 
This work connects the linguistic phenomenon with LLMs' computational mechanisms. We show how LLMs enforce certain computational mechanisms for the aforementioned tolerance in position perturbations.
Despite the complex design of the self-attention mechanism, this work reveals that LLMs learn a counterintuitive disentanglement of attention logits. Their values show a 0.959 linear correlation with an approximation of the arithmetic sum of positional relevance and semantic importance.
Furthermore, we identify a prevalent pattern in intermediate features, which we prove theoretically enables this effect. The pattern, which is different from how randomly initialized parameters would behave, suggests that it is a learned behavior rather than a natural result of the model architecture.
Based on these findings, we provide computational explanations and criteria for LLMs' position flexibilities. This work takes a pioneering step in linking \phenomenon{} with modern LLMs' internal mechanism.

%% file: figures/teaser.tex
\begin{figure}[t]
    \centering
    \includegraphics[width=0.48\textwidth]{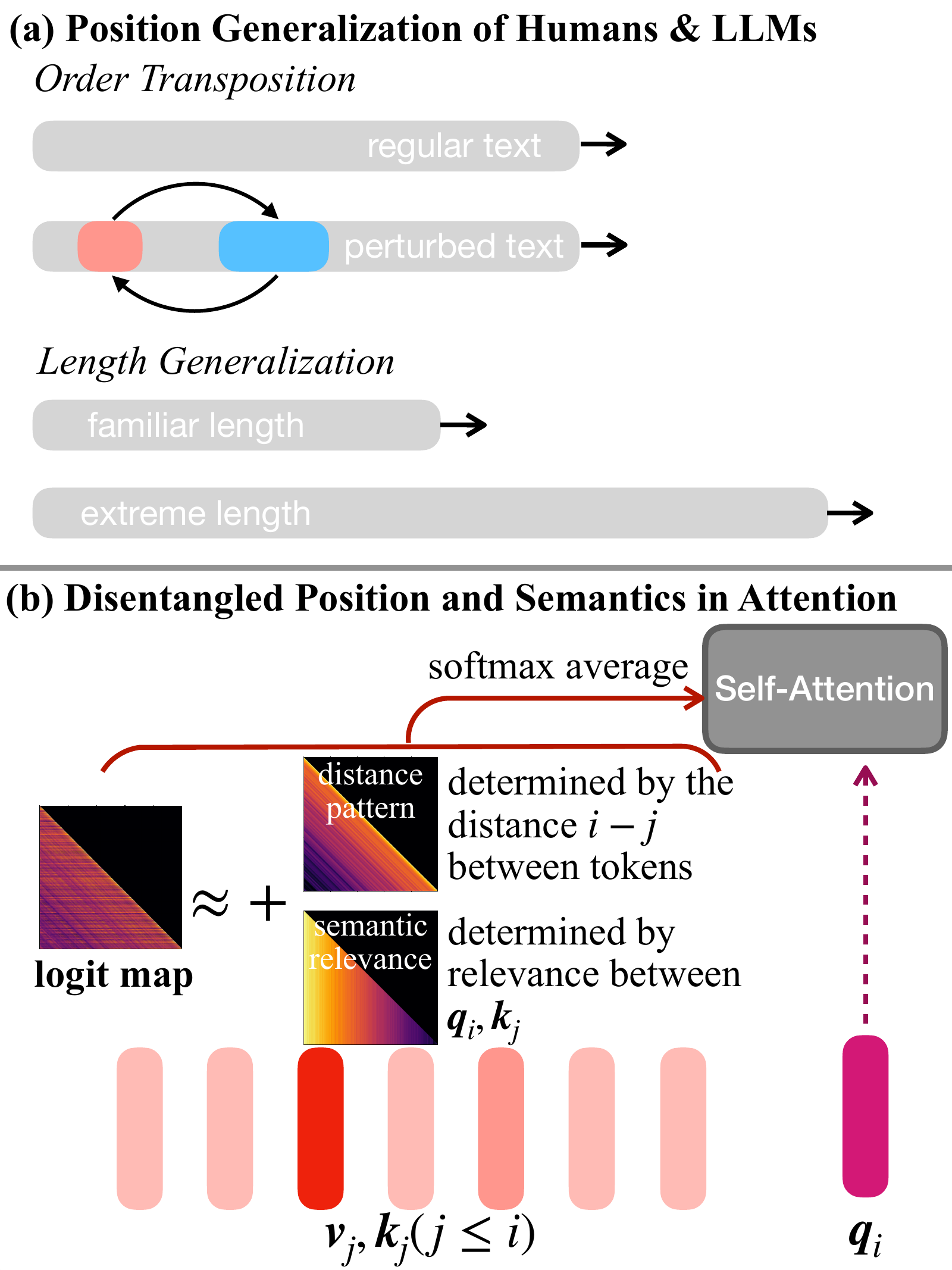}
    \caption{\textbf{(a)} LLMs, like humans, exhibit position generalization in various forms. \textbf{(b)} Self-attention in LLMs disentangles positional and semantic components so as not to be sensitive to position perturbations. The ``distance pattern'' and ``semantic relevance'' matrices show two subcomponents of the logit map that depend on positional and semantic relation, respectively.
    }
    \label{fig:teaser}
\vspace{-6mm}
\end{figure}

%% file: figures/logit_map.tex
\begin{figure*}[t]
    \centering
    \includegraphics[width=\textwidth]{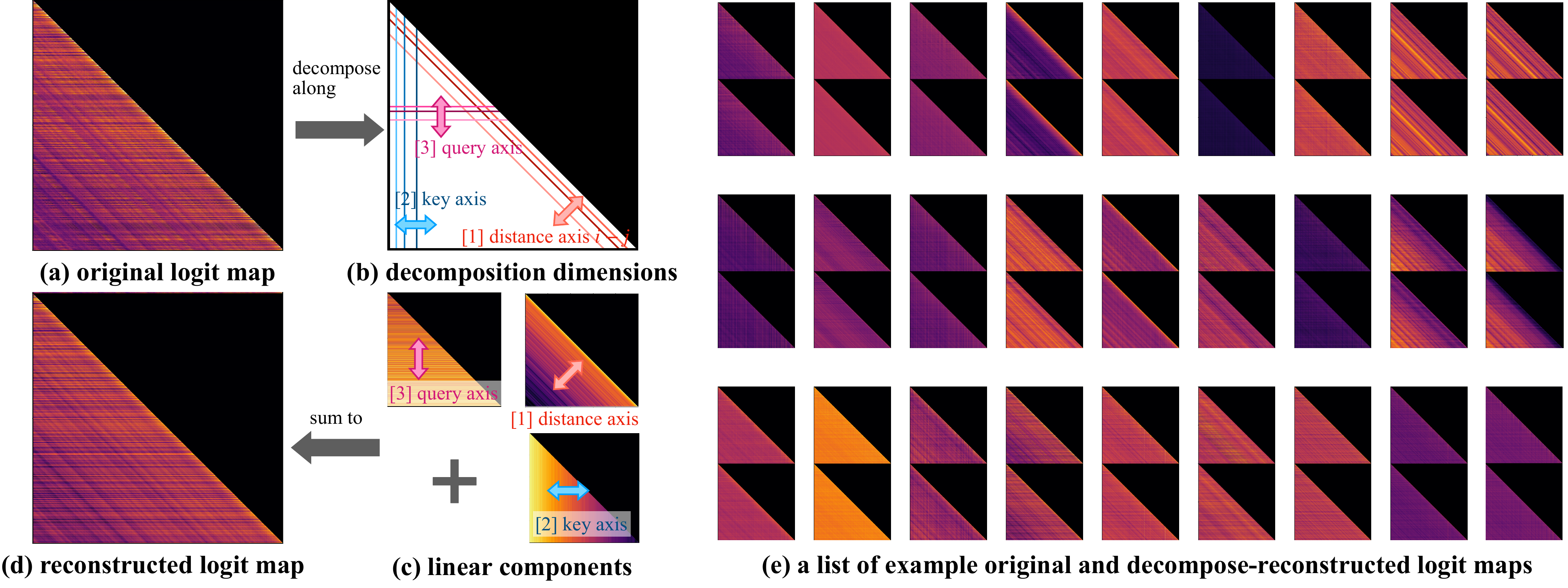}
    \caption{As a starting point of our study, we find that a 3-axis linear approximation (\textbf{(a)}$\rightarrow$\textbf{(d)}) is surprisingly similar to the original attention logit maps. Fig (e) is a set of original logit maps (upper ones) and their constructions (lower ones). More details are in Sec~\ref{subsec:real_attention}.}
    \label{fig:logit_map}
\vspace{-4mm}
\end{figure*}

%% file: body/1_introduction.tex
\section{Introduction}
\label{sec:introduction}

Most natural languages are written as sequences of textual elements such as characters, words, and sentences.
Despite this sequential nature, large language models (LLMs) exhibit remarkable tolerance in handling textual positions, just as observed in human studies~\cite{bruner1958note, rawlinson2007significance}. LLMs can comprehend text with position perturbations~\cite{sinha2021unnatural, pham2021out} and generalize to longer sequences than those seen during training with techniques like LM-Infinite~\cite{han2024lm} and InfLLM~\cite{xiao2024infllm}. These raise the question of how positional relevance is handled internally.
While prior research has explored various positional encoding strategies~\cite{su2021roformer, press2021train}, the underlying computational mechanisms of LLMs' position robustness remain largely unexplored.

In this work, we analyze the self-attention mechanism of modern LLMs to investigate how they process positional information to enable these capabilities. Our study reveals that LLMs learn a counter-intuitive \textit{disentanglement} in attention logits (Sec~\ref{subsec:real_attention}, \ref{subsec:fake_logits}). With a linear sum of two components $f(\vq, i-j) + g(\vq, \vk)$, which are about positional relation $i-j$ and semantical relation $g(\vq, \vk)$, respectively, the attention logits can be approximated with >0.95 linear correlation.
Furthermore, we identify a systematic pattern in intermediate representations, which we theoretically prove that enables this effect (Observation~\ref{obs:feature_pattern} and Theorem~\ref{thm:disentangle} in Sec~\ref{subsec:feature_pattern}). This pattern is different from how randomly initialized parameters of LMs would behave, which suggests that it is a learned behavior rather than an inherent consequence of model architecture.

Finally, we apply these findings to provide a computational explanation for the \phenomenon{} phenomenon in LLMs (Sec~\ref{sec:robustness}).
We demonstrate how text order transpositions on up to 5\% of all words only marginally affect the LLM's perplexity and downstream performance. This linguistic observation can be simulated by transposing the order of hidden features or perturbing the positional indices in relative position encoding, suggesting an analogy between human behaviors and the LLM computational mechanism.
We further explain how length generalization techniques can extend LLMs to extreme lengths without parameter updates. Taking insights from our analysis,
we show how self-attention is relatively tolerating while still ensuring the attention output vectors $\vo$ fall within the training-time distribution. This explains how feature distribution shift is avoided in length generalization techniques.

%% file: body/2_related.tex
\section{Related Work and Background}
\label{sec:related}

\subsection{Self-Attention and Positional Encoding}
\label{subsec:attention_background}
Self-attention is the core design in most modern LLMs for information flow to words from their contexts~\cite{vaswani2017attention}. It is also the primary (and often the only) component to inject text position information since the introduction of relative position encoding \cite{su2021roformer, touvron2023llama2, dubey2024llama, openai2023gpt4}, which is the subject of investigation in this work.
Despite architecture variants, it is generally designed as a Softmax-based weighted average over contextual ``value'' vectors $\{\vv_j| j \leq i \}$ before the current position $i$. The average weights $w( \vq_i, \vk_j, i-j)$ are determined by the relevance between the current word's ``query'' vector $\vq_i$, contextual words' ``key'' vectors $\{\vk_j|j\leq i\}$, and their relative position $i-j$. The output feature vector for the current token $\boldsymbol{o}_i$ is therefore:
\begin{equation}
\label{eq:attention}
    \vo_i =
    \sum_{j\leq i} \frac{w( \vq_i, \vk_j, i-j)}{\sum_{j'\leq i}\exp{w( \vq_i, \vk_{j'}, i-j')}} \boldsymbol{v}_j.
\end{equation}
Despite the existence of other choices of the function $w(\cdots)$ like Alibi~\cite{press2021train}, the de-facto mainstream choice is RoPE~\cite{su2021roformer}. It decomposes $\vq$ and $\vk$ vectors into 2-D tuples and lets them rotate in angle $(i-j)\theta_r$, where each 2-D tuple $r$ has a different rotating ``angular speed'' $\theta_r$.

\subsection{\Phenomenon{} (of both LLMs and Humans)}

Both humans and LLMs exhibit the ability to understand language with variable word or sentence positions. This phenomenon is related to multiple concepts from different perspectives.
Although written languages are usually represented as sequences of textual elements (such as characters, words, and sentences), they differ in their \textbf{Word Order Flexibility}~\citep{bakker1998flexibility, kaiser2004role}. Some languages (e.g., English, Chinese, Vietnamese, Indonesian) require a strict word order, while others (e.g., Hungarian, Japanese and Latin) allow more flexibility in order, 

which encodes pragmatic information such as emphasis~\cite{payne1992pragmatics}. 
This aspect has been computationally measured~\cite {kahane2023word} and used to evaluate linguistic complexity~\cite{szmrecsanyi2016informationtheoretic}.

Nevertheless, even when texts are perturbed to the extent that they no longer conform to regular language, humans can still understand them under certain conditions. The \textbf{Transposed Letter Effect}~\cite{bruner1958note, rawlinson2007significance} describes the ability to understand texts when the letter order is scrambled within words.
Language models also demonstrate the ability to perform downstream tasks on syntactically scrambled inputs, as shown in \textbf{Unnatural Language Processing}~\cite{sinha2021unnatural, pham2021out}. \citet{sinha2021masked} report comparable or improved quality of masked language models after pre-training on such corpora. At the sentence level, models pre-trained on randomly ordered corpora show improved performance on tasks involving complex contextual reasoning~\citep{shicontext}.

Preliminary studies on neural mechanisms underlying these phenomena in humans have been conducted in cognitive neuroscience~\cite {garcia2010transposition, dunabeitia2012differential, carreiras2015orthographic}, showing prevalent while varying robustness to transposition effects on letters, digits, and symbols in human brains. 

This work offers a computational counterpart, interpreting how \phenomenon{} is reflected in the internal mechanism of LLMs.

%% file: figures/logit_disentangle.tex
\begin{figure*}[t]
    \centering
    \includegraphics[width=\textwidth]{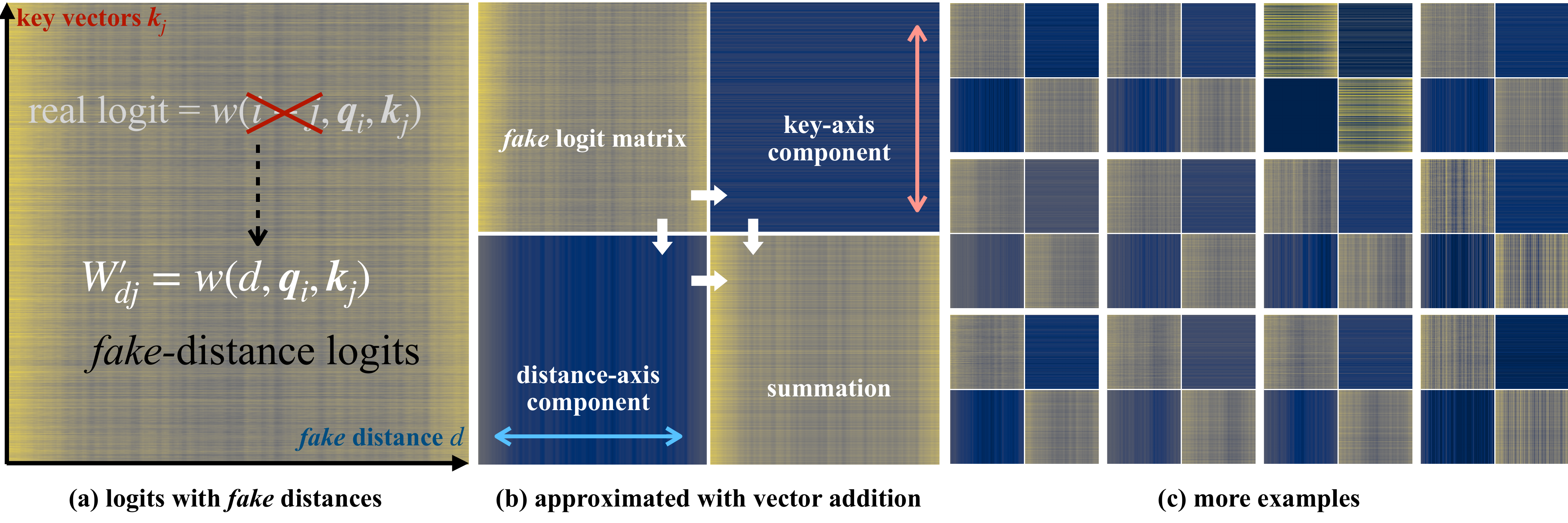}
    \caption{After replacing the distance value $i-j$ with a controlled fake distance $d$ (illustrated in \textbf{(a)}), we find that a distance-axis + key-axis decomposition closely resembles the logit calculation. \textbf{(b)} illustrates the disentanglement process. The key-axis and distance-axis components align well with the patterns of the \textit{fake}-distance logit matrix and sum up to a close approximation at the lower right corner.
    \textbf{(c)} presents additional examples, showing the prevalent applicability of such approximation.
    More details are provided in Sec.~\ref{subsec:fake_logits}.}
    \label{fig:logit_disentangle}
\end{figure*}

%% file: body/3_disentangle.tex
\section{LLMs Disentangle Position and Semantics in Attention}
\label{sec:disentangle}

How do LLMs handle the interaction between positional relation and semantic relation?
The attention logit function does not need to be smooth or simple across distances. It can be designed with arbitrary complexity so that at every distance $i-j$, the function $w(i-j, \cdot, \cdot)$ behaves drastically differently.
RoPE adopts a complex design that could theoretically implement (inverse) discrete Fourier transform, allowing it to approximate arbitrary functions with a sufficiently large dimension size. Unless otherwise stated, we use the Llama-3.2-7B model as the subject of study and extend results to other models in the Appendix.

Counter-intuitively, in this section, we reveal that LLMs learn a special feature pattern to empirically simplify the logit function $w(\cdots)$. The resulting attention logits can be approximately disentangled as an arithmetic addition of position relevance (determined by $i-j$ and $\vq_i$) and semantic importance (determined by $\vk_j$).
We will start with an interesting observation of low-rank components in attention maps in Sec~\ref{subsec:real_attention}.
Taking it as an inspiration, Sec~\ref{subsec:fake_logits} shows how the attention logits can be approximately disentangled into position and semantic-related components.
Sec~\ref{subsec:alibi} generalizes the discussion to the relative position encoding scheme in general.
Finally, Sec~\ref{subsec:feature_pattern} shows how LLMs computationally achieve this mechanism by enforcing a special pattern in key and query vectors.

%% file: body/3-1_real_attention.tex
\subsection{Starting Point: 3-Axis Linear Approximation of Logit Matrix}
\label{subsec:real_attention}

Let us start by looking at an attention head's logit matrix $W \in \gR^{n\times n}$ with elements $W_{i,j} = w( \vq_i, \vk_j, i-j)$ in Fig~\ref{fig:logit_map}(a). It is lower-triangular in causal language models where only past tokens are within the attention scope of the current token.
Despite combining information of three variables $i-j, \vq_i, \vk_j$, there are visible 1-d patterns along horizontal, vertical, and (off-)diagonal axes. These three axes are coincidentally the ones associated with the three variables as depicted in Fig~\ref{fig:logit_map}(b): in axis 1, $i-j$ does not vary in a diagonal line (the ``distance axis''); in axis 2, $\vk_j$ and $j$ do not vary in a vertical column (the ``key axis''); in axis 3, $\vq_i$ and $i$ do not vary in a horizontal row (the ``query axis'').

Inspired by this observation, we operate a ternary linear approximation of the logit map along the three axes. In other words, we examine if the logit map can be approximated with
\begin{equation}
\label{eq:ternary_logit_approx}
    W_{i,j} \approx a_{i-j} + b_i + c_j
\end{equation}
with three arrays (or three linear components) of variables $\va, \vb, \vc$.
To obtain an approximation, we formulate this as ridge regression, with more details in Appendix~\ref{sec:ternary_approximation_details}.
An example set of solved arrays is illustrated in Fig~\ref{fig:logit_map}(c).
\footnote{Note that this is different from a low-rank approximation of matrices, which approximates the full matrix (instead of the lower-trangular part), and only involves the row and column axes without diagonal components.}
By summing the three obtained components, the reconstructed logit matrix in Fig~\ref{fig:logit_map}(d) shows striking similarity with the original matrix. With more comparative examples shown in Fig~\ref{fig:logit_map}(e) and Appendix~\ref{sec:more_examples},
we show the prevalence of this trend across layers and attention heads.
This simple approximation has a correlation coefficient of 0.8650. This shows that a great majority of $W$'s variance can be explained by Eq~\ref{eq:ternary_logit_approx}. Interestingly, its linear nature implies the logit map contains simple components that vary only depending on key, query, or position information individually, but not their combinations.

This approximation, however, only serves as a starting point for our study, as the $\boldsymbol{a}$ component assumes a static and global positional pattern $a_{i-j}$ depending on token distance $i-j$, due to the course granularity of the analysis. We will move to a more fine-grained analysis in the next subsection.

%% file: figures/rotating_patterns.tex
\begin{figure*}[t]
    \centering
    \includegraphics[width=\textwidth]{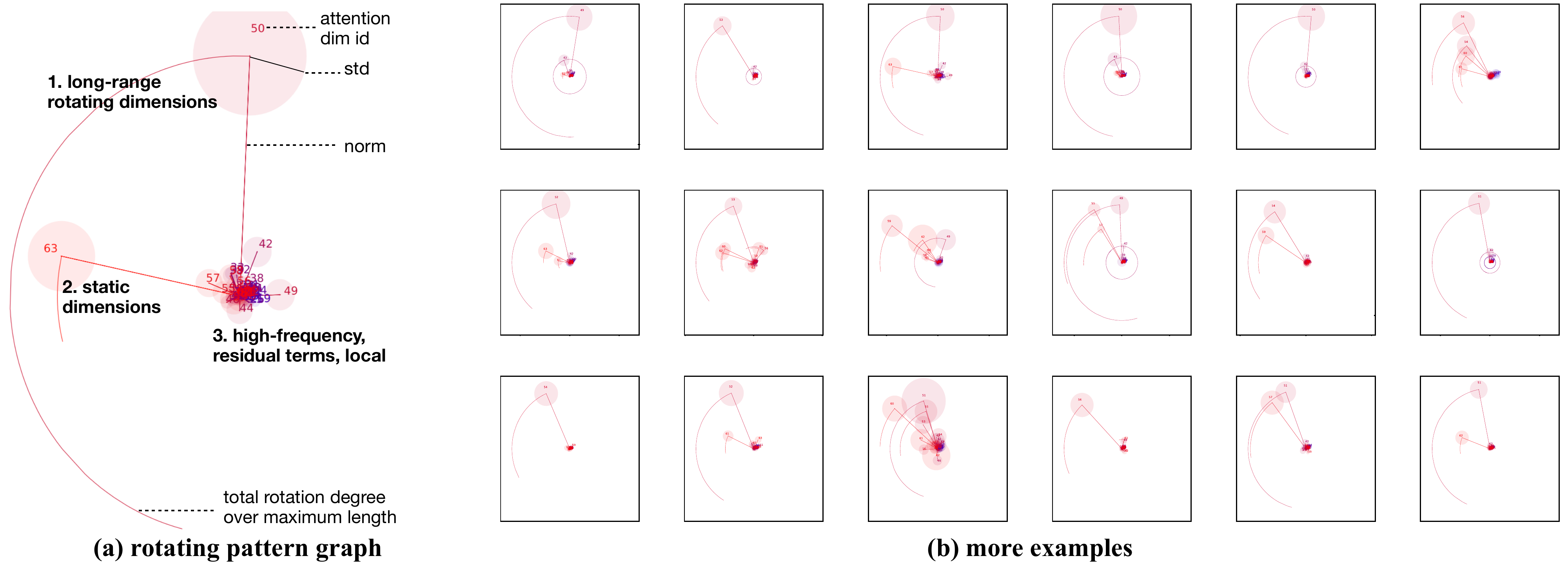}
    \caption{Visualization of rotating query-key vector tuples in RoPE-based attention described in Section~\ref{subsec:feature_pattern}. \textbf{(a)} The rotating tuples averaged over tokens are plotted as arrows, with tuple indices annotated from $0$ to $d/2-1$. Standard deviations over the tokens are shown as circles around endpoints, and the arc indicates the maximum rotation over the pre-training cutoff length. The sum of the tuples' projection along the horizontal axis is the actual logit value. \textbf{(b)} lists more of such figures, with more details in Sec~\ref{subsec:feature_pattern}}
    \label{fig:rotating_patterns}
\end{figure*}

%% file: body/3-2_attention_logits.tex
\subsection{The Disentanglement Law of Attention Logits}
\label{subsec:fake_logits}

The previous section identified independent linear patterns in the logit map. However, in the calculation of the attention logits $w(i-j, \vq_i, \boldsymbol{k}_j)$, the three variables still depend on each other. This prevents us from studying their effects on the logits individually.
Additionally, the ``query axis'' does not have an actual effect on LLMs. It applies a uniform offset on each logit row, which is also a uniform offset in Eq~\ref{eq:attention}. However, the softmax operator is invariant under uniform offsets.
\footnote{The softmax weights remain the same after an offset: $\frac{\exp (w_i+c)}{\sum_j \exp (w_j + c)}= \frac{\exp w_i\cdot \exp c}{\sum_j \exp w_j \cdot \exp c} = \frac{\exp w_i}{\sum_j \exp w_j}$.}
So, it would make more sense to control the query axis while fully disentangling the effect of the position and semantics axes.

Therefore, instead of studying the real attention logits, we use a \textit{fake} distance value $d$ to replace the real distance $i-j$: $w(d, \vq_i, \vk_j)$. In light of the previous discussion, we fix a query vector $\vq$ (for which we use the last token in the document to maximize the attention scope) and visualize the following fake logit matrix $W' \in \sR^{n\times n}$ where $W_{d,j}' = w(d, \vq_i, \vk_j)$ in Fig~\ref{fig:logit_disentangle}(a).
After this substitution, the new matrix $W'$ shows apparent vertical and horizontal patterns, suggesting prominent distance-wise and key-wise components.

We follow on disentangling $W'$ along these directions as  $ W_{d,j}' \approx a_d + b_j $.\footnote{This is a simplification of rank-2 matrix approximation.}
The least-square ridge regression solution of this approximation has an explicit-form solution (with more details in Appendix~\ref{sec:fake_logit_approximation}):
\begin{equation}
\begin{split}
    a_d & =  \frac{1}{n} \sum_{j'} W_{d, j'}' - \frac{1}{2n^2}\sum_{d', j'} W_{d', j'}' \\
    b_j &= \frac{1}{n}  \sum_{d'} W_{d', j}' - \frac{1}{2n^2}\sum_{d', j'} W_{d', j'}'. \\
\end{split}
\end{equation}
Essentially, $\va$ and $\vb$ are the average column and row of $W'$, respectively, with a constant offset of $- \frac{1}{2n^2}\sum_{d', j'} W_{d', j}'$.
This disentanglement process is visualized in Fig~\ref{fig:logit_disentangle}(b), where the key-axis and distance-axis components at two corners align with the patterns of the original \textit{fake} logit matrix well. Once we combine these two components, their summation again demonstrates high similarity with the original matrix with details.
This approximation has an average correlation coefficient of 0.9470 across layers, with a minimum linear correlation of the disentangling approximation of 0.914 and a maximum of 0.967, explaining the vast majority of the logits' variance by the two simple 1-dimensional components.
We list more examples of such approximation in Fig~\ref{fig:logit_disentangle}(c) and Appendix~\ref{sec:more_examples}.
These results indicate an approximated disentanglement of attention logits between positional relevance and semantic relevance:
\begin{equation}
\label{eq:logit_approximation}
    w(i-j, \vq, \vk) \approx f(\vq, i-j) + g(\vq, \vk)
\end{equation}.
In other words, the majority of contribution from the position relation of two tokens $f(\vq_i, i-j)$ is computed independently from their semantic relation $g(\vq_i, \vk_j)$ and added together.

\paragraph{Discussion: Triviality of Eq.~\ref{eq:logit_approximation}}
We here argue that RoPE does not inherently disentangle positional and semantic information by design.
As an empirical verification, we additionally experiment on a randomly initialized Llama-3.2-3B model. Using the same setup as described before, we carry out a 3-axis linear approximation of the logit map in this model on an input of length 4096. In this case, we only observe a 0.028 linear correlation coefficient between the original and approximated attention logits.
RoPE is mathematically equivalent to evaluating an inverse discrete Fourier transform along position $d$, which can express arbitrarily complex correlations between $\vq, \vk, d$ as shown in the following Theorem, with a proof listed in Appendix~\ref{sec:non_triviality}.

\begin{theorem}
\label{thm:non_triviality}
Let $f(d, x, y) \in [-C, C]$ be any bounded function over relative position $d \in \mathbb{Z}$ and semantic inputs $ x \in \mathcal{X}, y \in \mathcal{Y}$ with smoothness properties[6]. Then, for sufficiently large feature dimension $n=2N$, there exists a set of rotation frequencies $\{ \theta_r \} $, and features $ \boldsymbol{k}(x), \boldsymbol{q}(y) \in \mathbb{R}^n $ such that the RoPE attention logit approximates $ f(d, x, y) $ with arbitrarily small error:
\begin{equation}
\left| w(d, \boldsymbol{k}(x), \boldsymbol{q}(y)) - f(d, x, y) \right| < \epsilon.
\end{equation}
\end{theorem}

\subsection{Extending Discussions to General Relative Position Encoding}
\label{subsec:alibi}
This decomposition naturally extends to ALiBi~\cite{press2021train}, another most common form of positional encoding in current LLMs (e.g., MPT model family~\cite{MosaicML2023Introducing}). ALiBi introduces a position-dependent bias directly to the attention logits, which is linearly added and separable from the semantic interaction computed via the key-query dot product. This makes the position-semantic decomposition even more explicit and fits directly into the discussions above.

%% file: figures/position_perturbation_ppl.tex
\begin{figure*}[t]
    \centering
    \begin{subfigure}{0.32\textwidth}
        \centering
        \includegraphics[width=\textwidth]{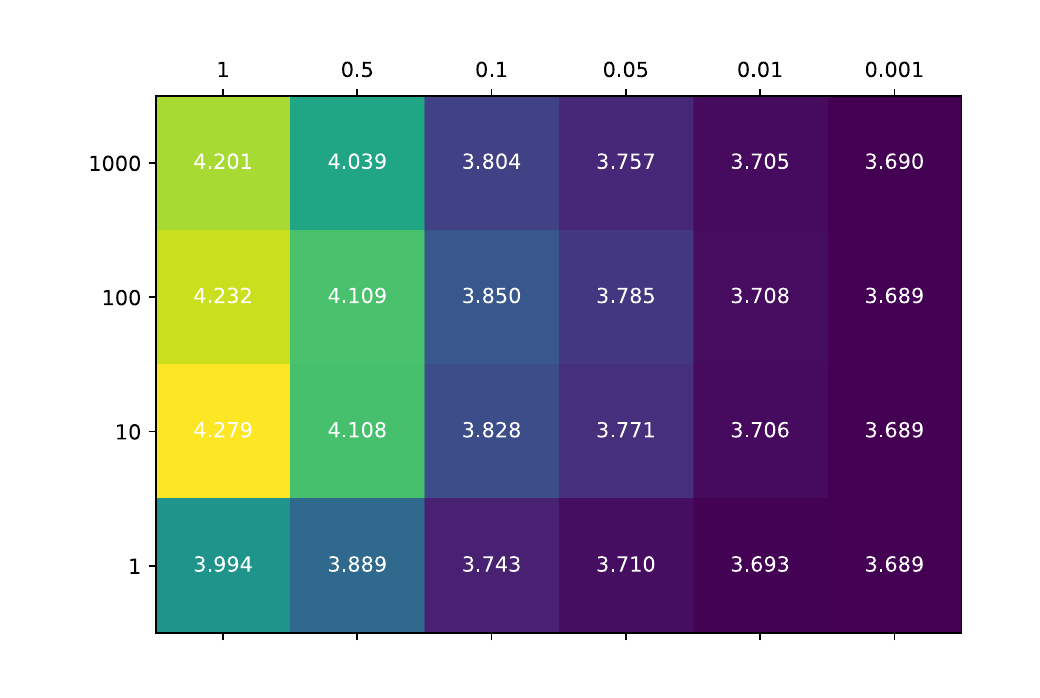}
        \caption{Effects of text transposition on LLM perplexity. $x$-axis controls the ratio of tokens perturbed, and $y$-axis controls the maximum distance of shuffled token pairs.}
        \label{fig:text_transposition}
    \end{subfigure}
    \hfill
    \begin{subfigure}{0.32\textwidth}
        \centering
        \includegraphics[width=\textwidth]{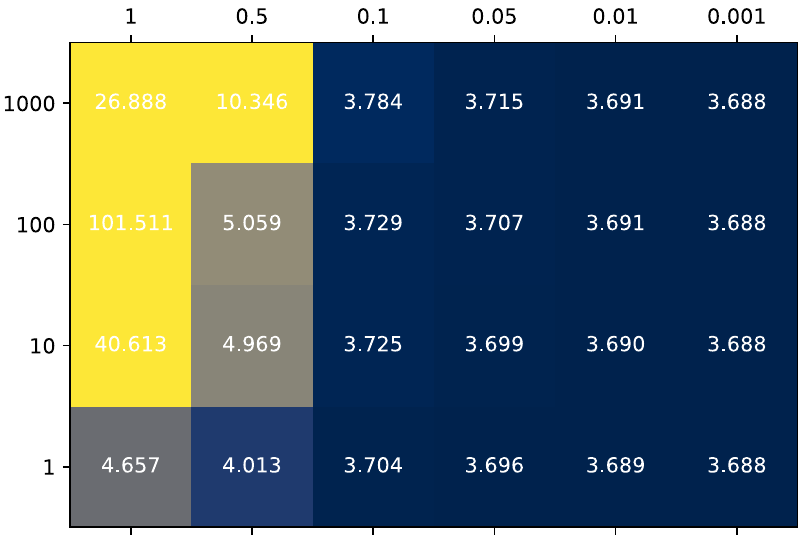}
        \caption{Effects of feature transposition on LLM perplexity. $x$-axis controls ratio of tokens with position indices perturbed, and $y$-axis controls the maximum value position offset.}
        \label{fig:feature_transposition}
    \end{subfigure}
    \hfill
    \begin{subfigure}{0.32\textwidth}
        \centering
        \includegraphics[width=\textwidth]{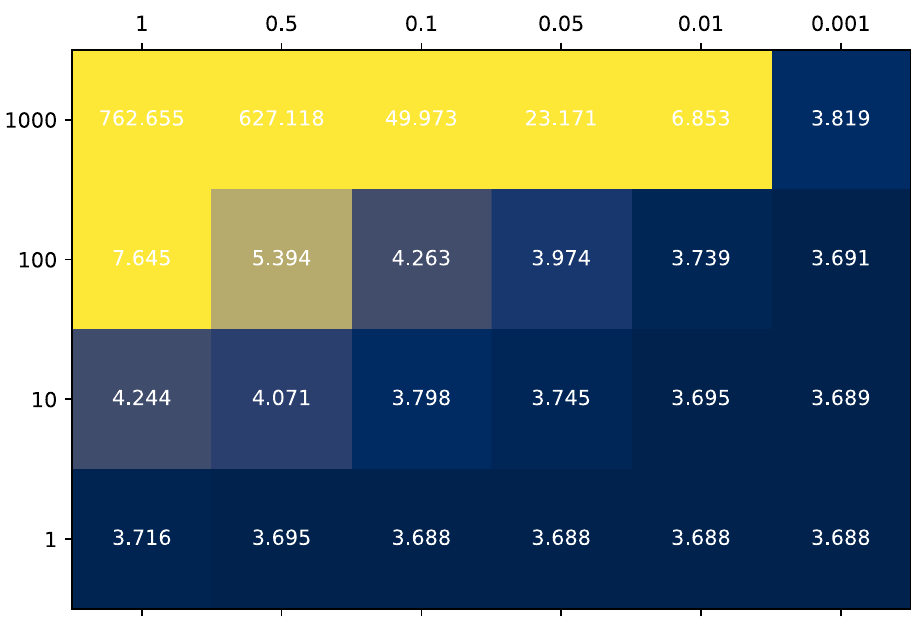}
        \caption{Effects of position encoding manipulation on LLM perplexity. $x$-axis controls ratio of tokens with position indices perturbed, and $y$-axis controls the maximum value position offset.}
        \label{fig:position_manipulation}
    \end{subfigure}
    \caption{Evaluating the impact of position information perturbation on LLMs' perplexity on ArXiv documents. With the vanilla perplexity being 3.688, our results show that shuffling text order in inputs and altering positional encodings in self-attention layers have limited effects on model perplexity and attention outputs.}
    \label{fig:position_perturbation}
    \vspace{-2mm}
\end{figure*}

%% file: body/3-4_feature_pattern.tex
\subsection{The Mechanism in Query-Key Space}
\label{subsec:feature_pattern}

What caused the phenomena mentioned in the last two sections?
Using the most prevalent positional encoding of RoPE as the subject of study,
we delve deeper into the hidden features to show that certain feature dimensions of $\vq$ and $\vk$ are enforced with a large fixed norm and direction so that the approximation in Eq~\ref{eq:logit_approximation} is possible.
Recall that the $d$-dimensional $\vq$ and $\vk$ are composed of a total number of $d/2$ 2-tuples rotating at different angular speeds, with lower-indexed tuples rotating much faster than high-indexed ones.
The overall logit
\begin{equation*}
\begin{split}
    &w(i-j, \vq, \vk) \\
    =& \sum_r \vk_r^\top M^{rot}_r((i-j)\theta_r) \vq_r \\
    =& \sum_r \|\vk_r\| \|\vq_r\|\cos\left((i-j)\theta_r + \theta_{\vq_r} - \theta_{\vk_r}\right)
\end{split}
\end{equation*},
which are sums of $\cos$ values of rotating vectors with norms of $\|\vk_r\| \|\vq_r\|$, starting angle of $\theta_{\vq_r} - \theta_{\vk_r}$ and rotating speed $\theta_r$ per distance.

We plot how these vectors would rotate together on a 2-D plane in Fig~\ref{fig:rotating_patterns}(a). The starting positions of these rotating vectors are plotted as arrows pointing from the point of origin. Each arrow's tuple index ($\in\{0 \ldots d/2-1\}$) is annotated beside the arrowheads.
To visualize the randomness in these vectors, we also plot their standard deviation as circles around the endpoints.
We also plot an arc to show the maximum rotation angle over the maximum distance allowed, i.e., the pre-training cutoff length $\theta_r^{\max{}} = \theta_r L_{\text{pre-train}}$.
Notably, there exist a few (two in the shown example) \textit{slow dominating} tuple dimensions with the following properties:

\begin{observation}
\label{obs:feature_pattern}
Properties observed in \textit{slow-dominating} features:
\begin{enumerate}
    \item (Prominent dimensions) A relatively fixed average starting vector $\E \vk_r$ with significantly larger norms than other dimensions that are not slowly rotating. In these slow-rotating dimensions, the deviation of vectors $\vk_r$ from the average vector $\E \vk_r$ is also small.
    \item (Dimensions that are mostly static) The total rotation angle $\theta_r^{\max}=\theta_rL_\text{pre-train}$ is usually small if the initial angle is close to $\pi$. 
\end{enumerate}
\end{observation}
More similar patterns can be found in Fig~\ref{fig:rotating_patterns}(b) and Appendix~\ref{sec:more_examples}.

We theoretically demonstrate how these patterns account for the previous entanglement in the sense that the contributions of \textit{slow dominating} tuple dimensions to logits disentangle the positional and semantic components. \textit{Other} tuple dimensions, however, contribute to relatively smaller variations in the logits.
We have the following asymptotic disentanglement of the logit function (with formal statements and proof in Appendix~\ref{sec:disentangle_theorem}):
\begin{theorem}
\label{thm:disentangle}
    There exists functions $f(\vq, i-j), g(\vq, \vk)$ that so that the effect of $i-j$ and $\vk$ can be asymptotically disentangled as:
    \begin{equation}
        w(i-j, \vq, \vk) = f(\vq, i-j)+g(\vq, \vk) + o(R)
    \end{equation},
    where $$R = \max\left(\text{Range}(f), \text{Range}(g) \right)$$ stands for the larger one of extreme range of $f$ and $g$ as $i, j, \vk$ vary
\end{theorem}.
Here, $f$ and $g$ are only related to the positional and semantic relation between tokens.
The logit function is approximated as the sum of two functions $f, g$, with a diminishing term compared to the function range of $f, g$.
This provides computational explanations for the observations in the previous two sections.
Not only are these functions existential, but the proof in Appendix~\ref{sec:disentangle_theorem} provides explicit-form solutions for $f, g$, which obtains a 0.959 linear correlation with the original logits.
This further validates the observations in Section~\ref{subsec:real_attention} and \ref{subsec:fake_logits}.

%% file: tables/qasper.tex
\begin{table}[h]
    \centering
    \setlength{\tabcolsep}{2pt}
    \begin{tabular}{p{2.5cm}ccccc}
        \toprule
        \textbf{Operation} & \multicolumn{5}{c}{\textbf{Qasper Accuracy}} \\
        \midrule
        & 0.5 & 0.1 & 0.05 & 0.01 & 0.001 \\
        \midrule
        \textbf{Original} & \multicolumn{5}{c}{42.53} \\
        \midrule
        \textbf{Text Order} & 37.39 & 41.44 & 42.34 & 42.37 & 42.53 \\
        \textbf{Feature Order} & 35.11 & 41.15 & 41.98 & 42.33 & 42.56\\
        \textbf{Position Encoding} & 37.19 & 42.44 & 42.42 & 42.74 & 42.64 \\
        \bottomrule
    \end{tabular}
    \caption{Effect of different levels of positional information perturbation on the Qasper Question-Answering dataset. Up to 5\% of the tokens can be transposed or applied with perturbed position encoding (within token distance $\pm 5$), while only resulting in a marginal effect on model accuracy.}
    \label{tab:qasper}
\end{table}

%% file: body/4_robustness.tex
\section{\Phenomenon{} of LLMs}
\label{sec:robustness}

Taking insights of findings in Sec~\ref{sec:disentangle}, this section explains how LLMs achieve \phenomenon{} towards \textit{perturbed text positions} and  \textit{unseen lengths}.
These phenomena reflect the aforementioned computational mechanism of disentangling position and semantics in attention: \textit{positional relevance is not tightly bonded with semantic information in attention inference}. Instead, they contribute linearly independently to attention logits.
In the following sections, we will empirically examine various forms of \phenomenon{} on the representation level.

%% file: tables/length_feature_kl.tex
\begin{table*}[ht]
\centering
\begin{tabular}{l|ccccc}
\toprule
\textbf{} & \textbf{4200} & \textbf{5120} & \textbf{6144} & \textbf{7168} & \textbf{8192} \\
\midrule
\textbf{KL (LM-Infinite)} & 0.0789 & 1.2093 & 2.6327 & 3.5192 & 4.4401 \\
\textbf{KL (Extending Vanilla LM)} & 0.5493 & 6.4862 & 8.6050 & 12.8479 & 15.9810 \\
\hline
\textbf{Log-Perplexity (LM-Infinite)} & 1.102 & 0.996 & 0.995 & 1.020 & 1.004 \\
\textbf{Log-Perplexity (Extending Vanilla LM)} & 1.316 & 7.268 & 8.737 & 8.754 & 8.901 \\
\bottomrule
\end{tabular}
\caption{KL divergence between the distribution of extra-length features and pretraining-length features is significantly smaller when a length extrapolation technique (such as LM-Infinite) is applied compared to vanilla LLMs. The distribution was approximated with a multivariant Gaussian distribution. This trend aligns with the lower loss (lower log-perplexity) when applying a length extrapolation method.}
\label{tab:length_table_kl}
\end{table*}

%% file: body/4-1_order.tex
\subsection{Tolerance to Position Perturbations}
\label{subsec:order}

Why can LLMs (like humans) read the language in shuffled word and sentence order?
In light of the analysis in Sec~\ref{sec:disentangle}, the positional information does not tightly bond with semantic relation but is more of an additive factor to the attention logit.
We experimentally examine how this mechanism affects the capability of LLMs on various levels. At the superficial level, we show that transposing the positions of a ratio of words in a text sequence has marginal effects on model behaviors. To investigate the reason further at the representation level, we also mimic the effect of text word perturbations in the LLM representations, such as shuffling the order of the feature sequences or modifying the position indices in positional encoding. More specifically:
\begin{enumerate}
    \item randomly shuffling a ratio $\gamma$ of the text orders in inputs within a maximum length $l_{\max}$ (\textbf{Text Transposition}),
    \item randomly shuffling a ratio $\gamma$ of the $\vk$ feature order in each attention layer within a maximum length $l_{\max}$ (\textbf{Feature Order Transposition}),
    \item randomly offsetting a ratio of $\gamma$ of the $\vk$'s position indices within self-attention layers within a range of $l_{\max}$ (\textbf{Position Encoding Manipulation})
\end{enumerate}.
We analyze the effects on attention output vectors and the corresponding LLMs' performance under these conditions.

The results, presented in Fig.~\ref{fig:position_perturbation}, show that LLMs exhibit robustness to these perturbation methods. The original model has a perplexity of 3.688 on the ArXiv documents~\cite{pile}.
Text transposition has a minimal impact on perplexity, with only a 0.02 increase when 1\% of tokens are shuffled up to a distance of 1000 tokens. This suggests that LLMs do not rigidly depend on strict word order.
Our intervention techniques simulate this linguistic transposition effect in Fig.~\ref{fig:feature_transposition} and ~\ref{fig:position_manipulation}.
Feature transposition also introduces a modest increase in perplexity, indicating that while position indices contribute to contextual representation, their precise ordering is not always critical in each self-attention layer. As further analysis, when the position encoding contains perturbed indices, the perplexity still has a marginal increase when 10\% of token positions were perturbed by $\pm 10$, or increase by an absolute value of 0.01 when 1\% of tokens has position encoding perturbed by $\pm 10$.
These phenomena further align with the previous observations that position information acts as a disentangled additive factor rather than being tightly entangled with semantic relationships.

As perplexity might not reflect a model's actual performance on downstream tasks, we evaluate how Llama-3.2-3B-Instruct performs on the Qasper dataset~\cite{Dasigi2021ADO} under these conditions. Results are listed in Table~\ref{tab:qasper}.
Similar to the findings on the model perplexity, the model can tolerate 5\% or word order being shuffled up to 5 token distance in the inputs, with only a 0.6\% drop in accuracy. When we perturb the positional information inside the model, the model exhibits flexibility (<0.1\% drop in accuracy) under both feature order transposition and position encoding manipulation when the position information of up to 10\% of the features is perturbed.

%% file: figures/sliding_window.tex
\begin{figure*}[t]
    \centering
    \includegraphics[width=0.9\textwidth]{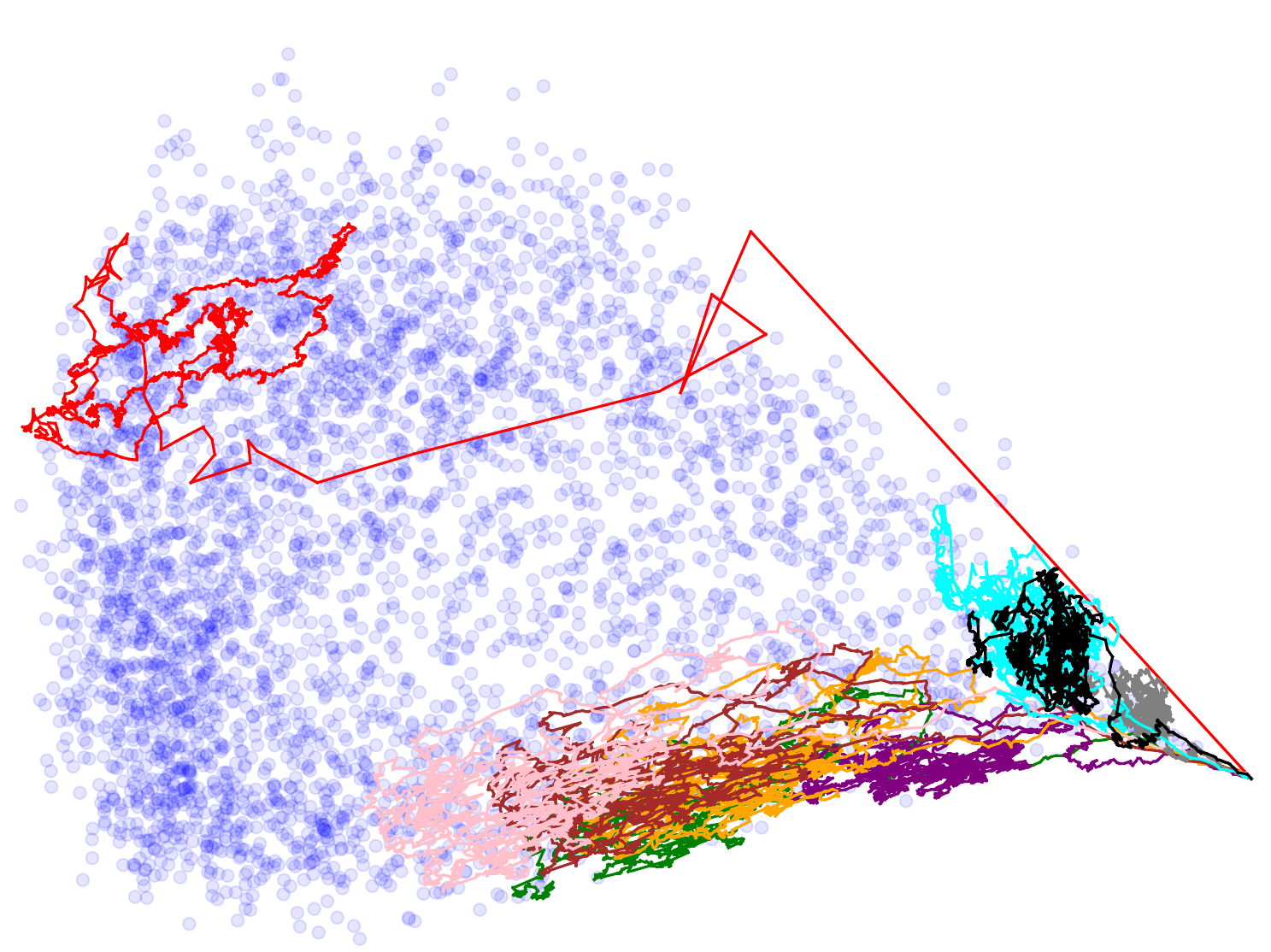}
    \caption{Visualization of attention output vectors projected onto a 2-D plane using PCA. Colored broken lines trace attention output vectors $\vo$ across a sliding window of key and value vectors in length extension techniques. This shows that the output vectors remain within the normal distribution, supporting our explanation of the possibility of length generalization.}
    \label{fig:sliding_window}
    \vspace{-4mm}
\end{figure*}

%% file: body/4-2_length.tex
\subsection{When and How LLMs Generalize to Longer Texts}
\label{subsec:length}
In vanilla relative position encoding, the list of distance values $[d]$ used in the attention operation between one query vector $q_i$ and all key features $k_j$ sequence is just the list of inter-token distances $[i, \ldots, 0]$. Recent techniques like LM-Infinite and InfLLM enable LLMs to generalize to longer text sequences than those encountered during training. The common practice adopted by these techniques is to modify the relative position before applying the original self-attention mechanism. This is equivalent to applying self-attention over a modified $[\vk_i]$ and $[\vv_i]$ sequence, which might be different (sometimes significantly shorter) than the original $[\vk_i]$ and $[\vv_i]$ sequences.
More specifically, in those techniques, the resulting sequence of features usually appears as if they are of the following positional distances:
\[
    [L_{PT}, \cdots, L_{PT}, L_{PT}-1, \cdots, l_L , \cdots, l_L, \cdots, 1, 0]
\],
where $L_{PT}$ is the pre-training maximum length, and $l_L$ is a position used technically for storing a few automatically retrieved feature vectors in the extremely long context. The retrieved features are usually used to enhance information retrieval.

This is in contrast to the intuition we obtained from common machine learning practices: why do LLMs train purely from shorter texts that generalize to extreme lengths (e.g., 200M in LM-Infinite) with only minor modifications to the model architecture? Moreover, little explicit design was implemented in modern SotA LLMs to enable this extreme generalization.
This phenomenon could find support in our analysis: \textit{even though posed to unseen extreme lengths, LLMs do not bind positional relevance information with the semantic features of the contextual tokens.} In other words, the $\vk$ and $\vv$ vectors could be interpreted as a pool of semantic features. The self-attention mechanism approximately and additively applies the position component to the pool. As long as the resulting distance list is similar to its normal shape $[L_{PT}, L_{PT}-1, \ldots,1, 0]$, the attention output vector will reside in its normal distribution. Therefore, technically, the attention output vector is still in-distribution, so the remaining parts of LLMs on top of the attention outputs will not take out-of-distribution features as inputs.

To verify this claim, we visualize the attention output vectors of an arbitrary layer using the technique above on the Pile dataset~\cite{gao2020pile}. This is a projection down to a 2-D plane using PCA\footnote{\url{https://scikit-learn.org/stable/modules/generated/sklearn.decomposition.PCA.html}}.
The blue dots are the normal attention output vectors, which mark their normal distribution.
Then, we select a set of $\vq$ vectors and associate them with different colors. For each vector, we apply it over a sub-sequence of length $L_{PT}$: $[\vk_i, \vk_{i+1},\ldots,\vk_{i+L_{PT}}]$ and $[\vv_i, \vv_{i+1},\ldots,\vv_{i+L_{PT}}]$. As we vary the value of starting position $i$, we trace the output vector with colored broken lines.
As shown in Figure~\ref{fig:sliding_window}, these lines, though extending to different directions and different ranges depending on the $\vq$, still wander within the range of normal attention output vector distribution.

In Table~\ref{tab:length_table_kl}, we also empirically verify that, as the context length increases, the distribution of length-generalizable text features remains similar to normal feature distributions, as shown in the following table. However, vanilla models on longer lengths will have an increasing KL divergence from normal features. This trend aligns with the trend of final perplexity.
This further validates our explanation of the length generalization and provides insights for future manipulation of the self-attention module for research purposes.

%% file: body/5_conclusions.tex
\section{Conclusions and Future Work}
\label{sec:conclusions}

In this work, we investigated the computational mechanisms behind the \phenomenon{} capabilities of LLMs. We first demonstrated that attention logits in LLMs can be approximately disentangled into independent components representing positional and semantic relevance. This finding suggests a structured decomposition within the model's internal computations.
Through empirical analysis, we further examined various forms of position generalization at the LLM representation level. These insights provide both computational explanations and insights into controlling these phenomena.

Future research could delve deeper into the specific architectural choices and training data patterns that contribute to this robustness. We also envision practical implications where the discovered decomposition might be important:
\begin{itemize}
    \item \textbf{Improved attention calculation}: after estimating the semantic importance and positional patterns, some (regions of) attention logits can be estimated without calculating precisely. This will be especially useful when those logits are low so that estimation errors will only translate to a small deviation in their exponentials. The insights offered in this work can also be used to inspire length extension methods of LMs, e.g., the explorations in later work \cite{bianchessi2025bayesian}.
    \item \textbf{KV cache compression}: Similar to the bullet above, the estimated importance component can be used as a precise and efficient criterion for filtering out $k, v$ caches that are more likely to receive low logits.
    \item \textbf{Position bias mitigation}: LLMs are reported to favor options that appear at specific positions~\cite{wang2024eliminating} or are lost in the middle~\cite{liu2023lost}. The discovered positional pattern and semantic importance can be used to re-calculate the attention logits, alleviating or removing such bias. 
\end{itemize}

%% file: body/x-1_limitations.tex
\section*{Limitations}
\label{sec:limitations}

While our study provides new insights into the computational mechanisms behind position generalization in LLMs, several limitations remain. First, our study primarily evaluates position robustness involving text order and length generalizations. While these are valuable computational linguistic phenomena, real-world language processing tasks often involve more complex positional dependencies, such as discourse coherence, document-level reasoning, and hierarchical structures. Future work could explore much more complicated scenarios. Second, our findings suggest that position and semantic components of attention logits can be disentangled, but the extent to which models actively leverage this property during training is unclear. Future explanations on how such a mechanism is acquired during training dynamics could greatly enhance the work.

%% file: body/x-2_acknowledgement.tex
\section*{Acknowledgement}

This research is supported by U.S. DARPA INCAS Program No. HR001121C0165, and DARPA ITM Program No. FA8650-23-C-7316. The views, opinions, and/or findings expressed are those of the authors and should not be interpreted as representing the official views or policies of the Department of Defense or the U.S. Government.

%% file: body/y-1_ternary_approximation.tex
\section{Details of Solving Ternary Linear Approximation in Sec~\ref{subsec:real_attention}}
\label{sec:ternary_approximation_details}

The total residue square in Eq~\ref{eq:ternary_logit_approx} is represented as:
\begin{equation}
    \mathcal{L}(W ; \boldsymbol{a}, \boldsymbol{b}, \boldsymbol{c}) = \sum_{j\leq i}(W_{i,j}-a_{i-j} - b_i - c_j)^2
\end{equation}.
This is a strictly convex function, so one single optimal solution exists. At the optimum, the objective has zero gradient $\nabla \mathcal{L}(W ; \boldsymbol{a}, \boldsymbol{b}, \boldsymbol{c}) = \boldsymbol{0}$. Taking derivative over all variables, this requirement is equivalent to a linear system:
\begin{equation}
\begin{split}
    (n-i)a_i + \sum_{j\leq n-i}b_j + \sum_{j\geq i}c_j = \sum_{j\geq i} W_{i+j, j} \\
    \sum_{j\leq n-i}a_j + (n-i)b_i + \sum_{j\geq i}c_j = \sum_{j\geq i} W_{j, i} \\
    \sum_{j\leq i} (a_j + b_j) + (i+1)c_i = \sum_{j\leq i}W_{i, j}
\end{split}
\end{equation}.
We then apply a linear equation solver\footnote{Adopted solver in NumPy: \url{https://numpy.org/doc/2.2/reference/generated/numpy.linalg.solve.html}. We add a 1e-6 $l_2$-norm regularization for numerical stability of solutions} to this system.

%% file: body/y-2_fake_logit_approximation.tex
\section{Details of the Approximation in Sec~\ref{subsec:fake_logits}}
\label{sec:fake_logit_approximation}

The total residue square in Eq~\ref{eq:ternary_logit_approx} is represented as:
\begin{equation}
    \mathcal{L}(W' ; \boldsymbol{a}, \boldsymbol{b}) = \sum_{d,j}(W_{d,j}'-a_d - b_j)^2
\end{equation}.
Taking derivative over all variables, the optimal point satisfies a linear system:
\begin{equation}
\begin{split}
    na_d + \sum_{j'} b_{j'} &= \sum_{j'} W_{d, j'}' \\
    \sum_{d'} a_{d'} + nb_j &= \sum_{d'} W_{d', j}' \\
\end{split}
\end{equation}.
The solution set is the following family of values, where $c$ can take arbitrary values.

\begin{equation}
\begin{split}
    a_d & =  \frac{1}{n} \sum_{j'} W_{d, j'}' + c -  \frac{1}{n^2}\sum_{d', j'} W_{d', j'}' \\
    b_j &= \frac{1}{n}  \sum_{d'} W_{d', j}' - c \\
\end{split}
\end{equation}.
Adding an $l_2$-norm regularization term with any weight, as $c$ is the only free variable here, the optimal solution will be the point where.
\begin{equation}
    \sum_d a_d - \sum_j b_j = 0
\end{equation}.
That will require $c=\frac{1}{2n^2}\sum_{d', j'} W_{d', j'}'$. The final solution will become:
\begin{equation}
\begin{split}
    a_d & =  \frac{1}{n} \sum_{j'} W_{d, j'}' - \frac{1}{2n^2}\sum_{d', j'} W_{d', j'}' \\
    b_j &= \frac{1}{n}  \sum_{d'} W_{d', j}' - \frac{1}{2n^2}\sum_{d', j'} W_{d', j'}' \\
\end{split}
\end{equation}.

%% file: body/y-3_pattern_mechanism.tex
\section{Asymptotic Disentanglement of Attention Logit Function}
\label{sec:disentangle_theorem}

\begin{assumption}
\label{asm:feature_pattern}
Consider a sequence of feature-related variables 
\(\{ \vk_r^{(n)}, \vq_r^{(n)}, \theta_r^{(n)}, L_{\text{pre-train}}^{(n)} \}_{n \in \mathbb{N}}\), 
where \( n \) represents an increasing parameter (e.g., as training steps evolve, which reflects the observation that the following observations are a learned behavior on pre-training data). 
For readability, we omit the explicit sequence index \( (n) \) in the following statements, 
but all asymptotic relations are understood to hold as \( n \to \infty \).
Properties observed in \textit{slow-dominating} features:
\begin{enumerate}
    \item (Prominent dimensions) A relatively fixed average starting vector $\E \vk_r$ with significantly larger norms than other dimensions. $\exists \gR_\text{slow}$ $\forall r'\not\in \gR_\text{slow}, r\in \gR_\text{slow}, \|\vk_{r'}\|\|\vq_{r'}\| = o(\|\vk_r\|\|\vq_r\|)$.
    Also $\forall r\in \gR_\text{slow}, \|\vk_r - \E \vk_r\| = o(\|\E \vk_r\|)$.
    \item (Dimensions that are mostly static) The total rotation angle $\theta_r^{\max}=\theta_rL_\text{pre-train}$ is usually small if initial angle is close to $\pi$, i.e, $\theta_r L_\text{pre-train} = o(\theta_{\vq_r} - \theta_{\vk_r} - \pi )$.
\end{enumerate}
\end{assumption}

Based on the assumptions above, we provide a more formal statement of Theorem~\ref{thm:disentangle} as follows:

\begin{theorem}
\label{thm:disentangle_formal}
    If feature properties described in Observations~\ref{asm:feature_pattern} holds, then there exists functions $f(\vq, i-j), g(\vq, \vk)$ that so that the effect of $i-j$ and $\vk$ can be asymptotically disentangled as:
    \begin{equation}
        w(i-j, \vq, \vk) = f(\vq, i-j)+g(\vq, \vk) + o(R)
    \end{equation},
    where $$R = \max\left(\text{Range}(f), \text{Range}(g) \right)$$ stands for the larger one of extreme range of $f$ and $g$ as $i, j, \vk$ vary.
\end{theorem}.

\begin{proof}

In those slow-dominating dimensions $r\in\gR_\text{slow}$, denote $\theta_{\delta}=\theta_{\vq_r}- \theta_{\vk_r} - \pi$ and $\bar{\vk} = \E \vk$.
Let $\text{Range}_x(f) = \sup_x(f)-\inf_x(f)$ denote the extreme range of function $f$ over a variable $x$ (out of potentially multiple variables).
We have:
\begin{equation*}
\begin{split}
    &\vk_r^\top M^{rot}_r((i-j)\theta_r) \vq_r \\
    = & \|\vk_r\|\|\vq_r\|\cos(\theta_{\vq_r}- \theta_{\vk_r} + (i-j)\theta_r)\\
    = & \|\bar{\vk_r} + (\vk_r - \bar{\vk_r})\|\|\vq_r\| \\
    & \cos(\pi + \theta_{\delta} + (i-j)\theta_r) \\
    \leq & - (\|\bar{\vk_r}\| + \|\vk_r - \bar{\vk_r}\|)\|\vq_r\| \\
    & \cos(\theta_{\delta} + (i-j)\theta_r) \\
    = & - \|\bar{\vk_r}\| \|\vq_r\| \cos(\theta_{\delta} + (i-j)\theta_r) \\
    & - \|\vk_r - \bar{\vk_r}\|\|\vq_r\| \\
    & \left( \cos\theta_{\delta} - 2\sin{(\theta_{\delta}+\frac{1}{2}(i-j)\theta_r)} \sin{\frac{1}{2}(i-j)\theta_r} \right)\\
    = & - \|\bar{\vk_r}\| \|\vq_r\| \cos(\theta_{\delta} + (i-j)\theta_r \\
    & + \|\vk_r - \bar{\vk_r}\|\|\vq_r\| \cos\theta_{\delta}) \\
    & + \|\vk_r - \bar{\vk_r}\|\|\vq_r\|\\
    & 2\sin{(\theta_{\delta}+\frac{1}{2}(i-j)\theta_r)}\sin{\frac{1}{2}(i-j)\theta_r}
\end{split}
\end{equation*}

Let
\begin{equation*}
\begin{split}
f_r(\vq_r, i-j) =& - \|\bar{\vk_r}\| \|\vq_r\| \cos(\theta_{\delta} + (i-j)\theta_r) \\
g_r(\vq_r, \vk_r) =& - \|\vk_r - \bar{\vk_r}\|\|\vq_r\| \cos\theta_{\delta} \\
l_r(\vq_r, \vk_r, i-j) =& - 2\|\vk_r - \bar{\vk_r}\|\|\vq_r\|\\
& \sin{(\theta_{\delta}+\frac{1}{2}(i-j)\theta_r)}\\
& \sin{\frac{1}{2}(i-j)\theta_r}  \\
\end{split}
\end{equation*}
.

Now let's split the case for discussion. When $\theta_\delta < \frac{\pi}{4}, \cos(\theta_\delta)\geq\frac{\sqrt{2}}{2}$, $(i-j)\theta_r = o(\frac{\pi}{4})$, 

\begin{equation*}
\begin{split}
    |l(\vq_r, \vk_r, i-j)| \leq& 2\|\vk_r - \bar{\vk_r}\|\|\vq_r\|\sin{\frac{1}{2}(i-j)\theta_r} \\
    =& o(\|\vk_r - \bar{\vk_r}\|\|\vq_r\|) \\
    =& o(\text{Range}_{\vk_r}(g_r(\vq_r, \vk_r)))
\end{split}
\end{equation*}

When $\theta_\delta \geq \pi / 4$, 
\[    \text{Range}(\cos(\theta_{\delta} + (i-j)\theta_r)) \geq \frac{1}{2}(\theta_rL_{pre-train})^2 \]

\begin{equation*}
\begin{split}
    |l_r(\vk_r, \vq_r, i-j)| \leq & 2\|\vk_r - \bar{\vk_r}\|\|\vq_r\|\sin{\frac{1}{2}(i-j)\theta_r} \\
    \leq & 2\|\vk_r - \bar{\vk_r}\|\|\vq_r\| \\
    =& o(\|\bar{\vk_r}\| \|\vq_r\|) \\
    =& o(\text{Range}_{i-j}(f_r(\vq_r, i-j)))
\end{split}
\end{equation*}

In summary, 
\begin{equation*}
\begin{split}
    \vk_r^\top M^{rot}_r((i-j)\theta_r) \vq_r =& f_r(\vq_r, i-j) + g_r(\vq_r, \vk_r)) \\
    &+ o((\text{Range}_{\vk_r}(g_r(\vq_r, \vk_r)) \\
    &+ \text{Range}_{i-j}(f_r(\vq_r, i-j)))
\end{split}
\end{equation*}

Then, the faster-rotating dimensions have contributions smaller than slower ones:
\begin{equation}
\begin{split}
    & \sum_{r\notin\gR_\text{slow}} \vk_r^\top M^{rot}_r((i-j)\theta_r) \vq_r) \\
    (r_0\in\gR_\text{slow})= & |\gR| o(\|\vk_{r_0}\|\|\vq_{r_0}\|) \\
    = & o\left(\sum_{r\in\gR_\text{slow}} \vk_r^\top M^{rot}_r((i-j)\theta_r) \vq_r\right)
\end{split}
\end{equation}

In summary:

\begin{equation}
\begin{split}
    & w(i-j, \vq, \vk) \\
    = & - \sum_{r\in\gR_\text{slow}}(\|\bar{\vk_r}\| \|\vq_r\| \cos(\theta_{\delta} + (i-j)\theta_r) \\
    & + \|\vk_r - \bar{\vk_r}\|\|\vq_r\| \cos\theta_{\delta}) (1+o(1)) \\
    = & f(\vq, i-j)+g(\vq, \vk) + \\
    & +  o((\text{Range}_{\vk}(g(\vq, \vk)) + \text{Range}_{i-j}(f(\vq, i-j)))
\end{split}
\end{equation}
if we define
\begin{equation}
\begin{split}
    f(\vq, i-j) &= -\sum_{r\in\gR_\text{slow}} f_r(\vq_r,i-j) \\
    g(\vq, \vk) &= -\sum_{r\in\gR_\text{slow}} g_r(\vq_r, \vk_r)
\end{split}
\end{equation},
respectively.

\end{proof}

%% file: body/y-5_more_examples.tex
\section{More Example Visualization from Other Models}
\label{sec:more_examples}

\begin{figure*}[t]
    \centering
    \includegraphics[width=\textwidth]{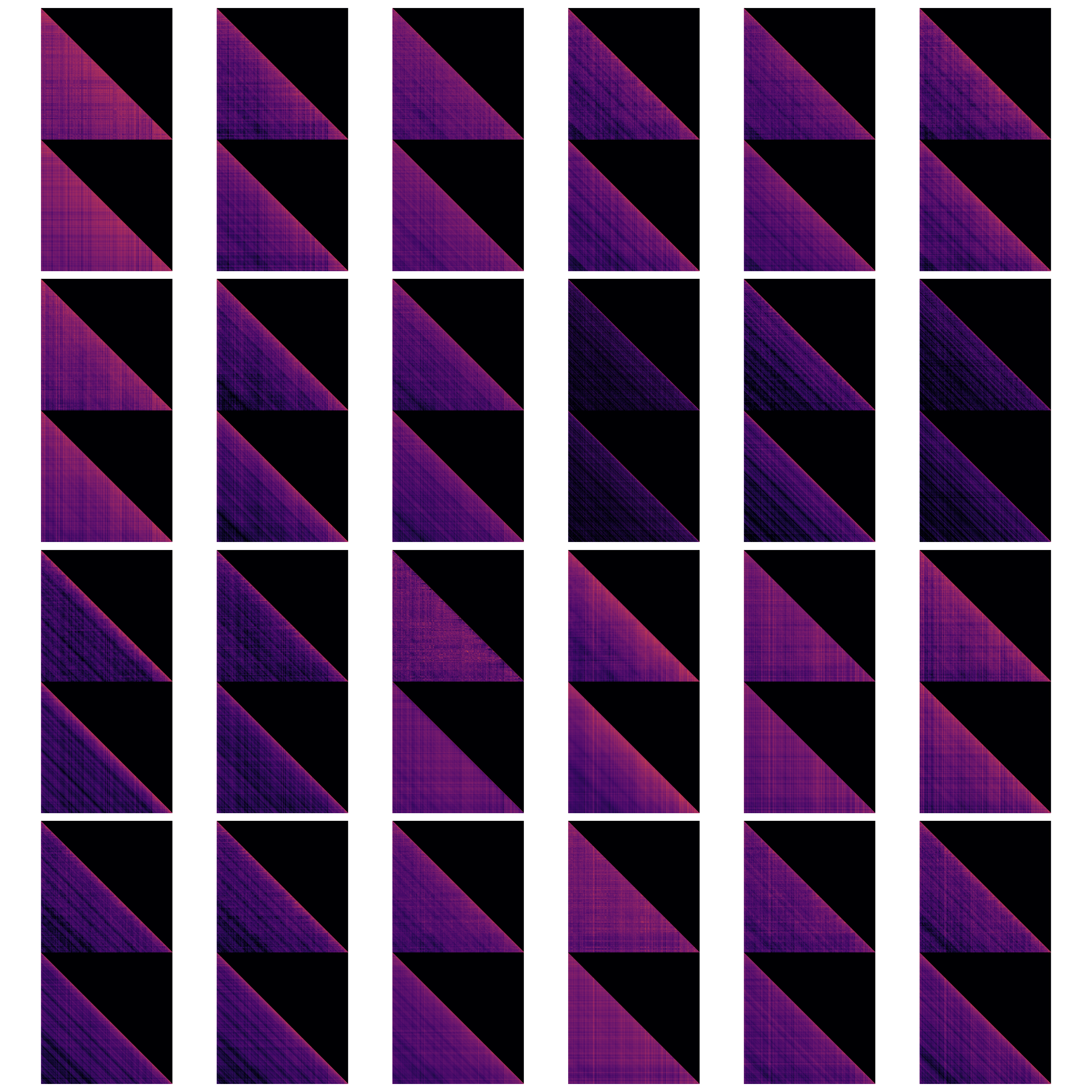}
    \caption{More examples of 3-axis approximation of logit matrix on Llama-2 model.}
\end{figure*}

\begin{figure*}[t]
    \centering
    \includegraphics[width=\textwidth]{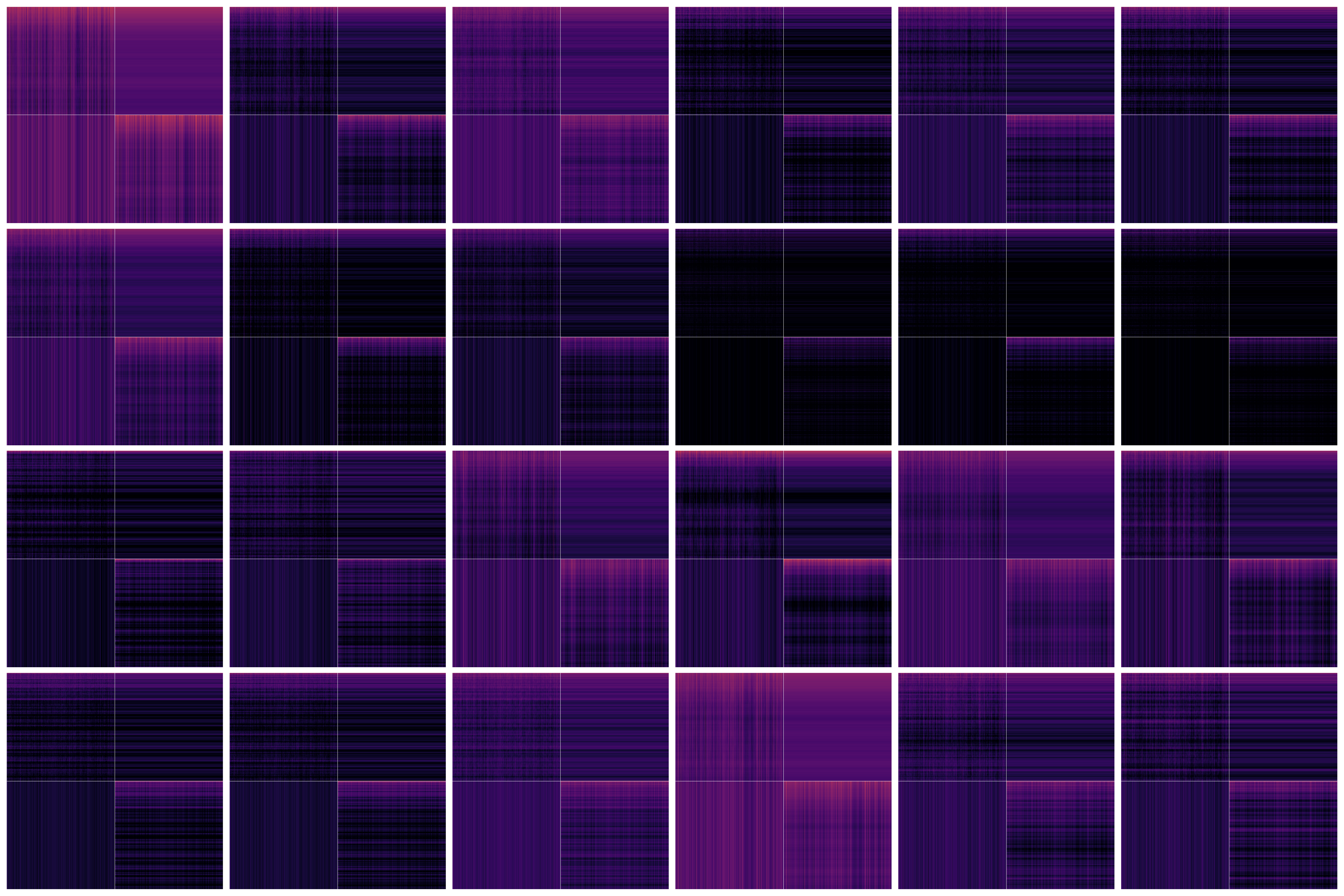}
    \caption{More examples of disentanglement of fake-distance logit matrix on Llama-2 model.}
\end{figure*}

\begin{figure*}[t]
    \centering
    \begin{subfigure}{\textwidth}
        \centering
        \includegraphics[width=\textwidth]{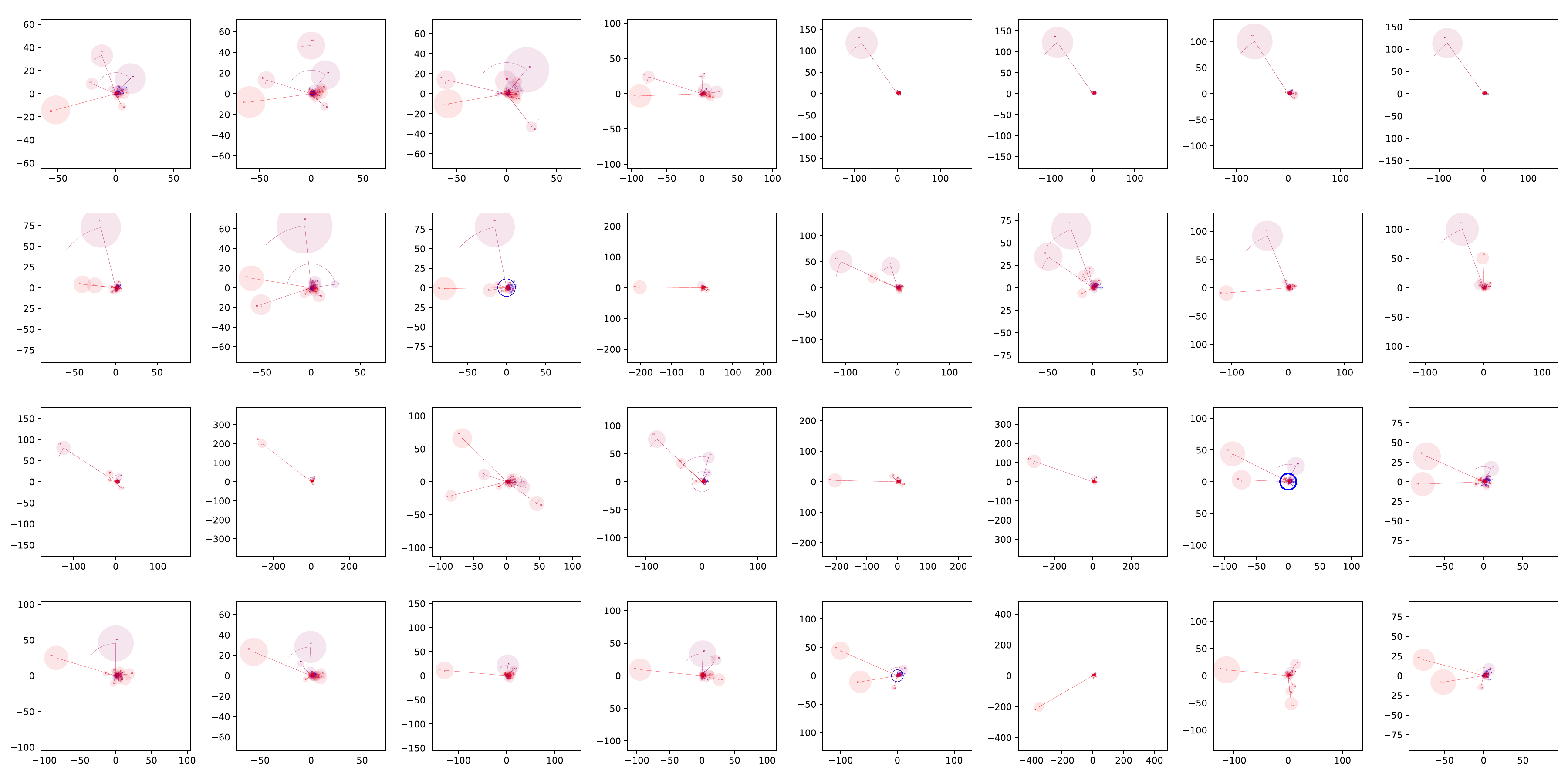}
        \caption{Llama-2.}
    \end{subfigure}
    \hfill
    \begin{subfigure}{\textwidth}
        \centering
        \includegraphics[width=\textwidth]{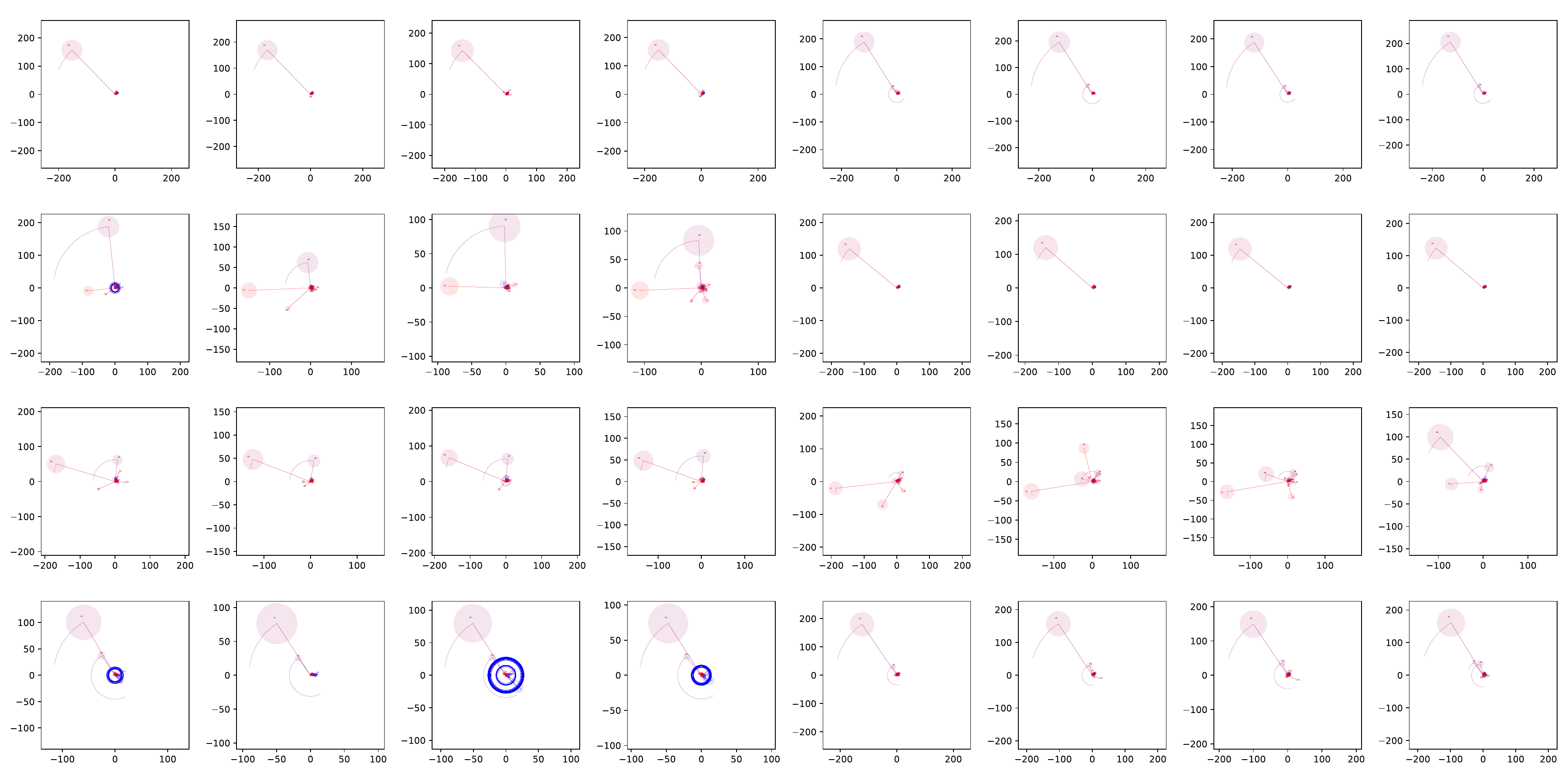}
        \caption{Llama-3}
    \end{subfigure}
    \caption{More examples of the rotating vector tuples in RoPE-based attention in other models.}
\end{figure*}

%% file: body/y-6_non_triviality.tex
\section{Proof of Theorem~\ref{thm:non_triviality}}
\label{sec:non_triviality}

\begin{proof}
Let $\theta_r=2\pi\frac{r}{N}$, and let us view $ f(d, x, y) $ along the variable of $d\in[0 ... L]$, and define the Fourier spectrum $F(x, y) = [F_0(x, y), ..., F_{N-1}(x, y)]$:
$$
 F_r(x, y):= \frac{1}{N} \sum_{d=0}^L f(d, x, y) e^{-\mathrm{i} d \theta_r}
$$

We now encode the spectrum into $ \boldsymbol{k}(x) $ so that
$$
\mathrm{i}k_{2r}(x) + k_{2r+1}(x) = \frac{F_r(x, y)}{\mathrm{i}q_{2r}(y) + q_{2r+1}(y)}
$$

Then we have:
\begin{align*}
&w(d, \boldsymbol{k}(x), \boldsymbol{q}(y)) \\
=& \sum_{r=1}^N \text{Re} \left[ F_r(x, y) e^{i d \theta_r} \right] \\
=& f(d, x, y) + O(e^{-cn})
\end{align*}
for some constant $c$ by established bounds of discrete (inverse) Fourier coefficients in \cite{nissila2018fourier}.
\end{proof}